\documentclass[10pt,conference]{IEEEtran}

\usepackage{amsmath,amssymb,amsfonts}
\usepackage{booktabs}
\usepackage{multirow}
\usepackage{graphicx}
\usepackage{xcolor}
\usepackage{url}
\usepackage{cite}
\usepackage{algorithm}
\usepackage{algpseudocode}
\usepackage{tikz}
\usetikzlibrary{arrows.meta,positioning,shapes.geometric,fit}
\usepackage{balance}

\newcommand{\method}{\textsc{DRAP}}
\newcommand{\kappatune}{KappaTune}

\newcommand{\eps}{\varepsilon}

\title{DRAP: Dynamics-Risk Adaptive Plasticity for Continual Learning in Physical AI}

\author{\IEEEauthorblockN{Oswaldo Ludwig}
\IEEEauthorblockA{\\
Email: oswaldoludwig@gmail.com}}

\begin{document}
\maketitle

\begin{abstract}
Continual learning in physical AI differs from conventional continual learning because small per-step policy errors can accumulate over a trajectory. Each deviation may move the agent outside the training distribution, increase subsequent prediction errors, and ultimately cause closed-loop task failure. This work introduces Dynamics-Risk Adaptive Plasticity (\method{}), a replay-free method that estimates the historical importance of adaptable model tensors from sample-level action gradients weighted by local trajectory-amplification risk. A ridge-regularized local linear model is fitted to expert state-action transitions, and the largest singular value of its state-transition matrix provides a policy-independent, Lyapunov-inspired estimate of one-step perturbation amplification. After each task, risk-weighted squared gradients are accumulated in a tensor-level importance memory. During subsequent tasks, \method{} uses normalized cumulative importance to scale the effective AdamW learning rate of each tensor, immediately reducing the plasticity of components that supported previous behavior at dynamically sensitive trajectory points. We instantiate \method{} on a fixed \kappatune-selected subnetwork of OpenVLA and evaluate replay-free sequential adaptation across the LIBERO Spatial, Object, and Goal suites. Compared with standard AdamW fine-tuning and EWC, the results show substantially improved retention of previously learned behaviors while preserving strong new-suite acquisition.
\end{abstract}


\section{Introduction}
Continual learning in physical AI differs from conventional continual learning because small changes in a policy can alter future observations and amplify into closed-loop task failure. In embodied control, an action error changes the next physical state, modifies the next observation, and may place the policy outside the training  distribution. Subsequent errors can therefore compound over the remainder of the trajectory \cite{ross2010efficient,ross2011dagger}.

Vision-language-action (VLA) models combine visual perception, language grounding, and low-level control within a single pretrained policy. OpenVLA provides an open 7B VLA with tokenized actions and a pretrained vision-language backbone \cite{kim2024openvla}. OpenVLA-OFT later demonstrated that continuous actions, action chunking, and optimized decoding can substantially improve success and inference efficiency \cite{kim2025oft}. These advances strengthen task acquisition, but do not remove the continual-learning problem: sequential fine-tuning can overwrite behavior required by earlier manipulation skills.

Existing continual-learning strategies offer complementary trade-offs. Replay remains a strong baseline but requires storing or regenerating previous observations. Expansion and routing methods isolate tasks but add modules or routing mechanisms. Parameter-regularization methods preserve a previous solution but can restrict acquisition. Direct update-suppression methods reduce future plasticity according to past update history, but can become too restrictive for large heterogeneous models.

This work develops a replay-free mechanism based on a physical observation: parameters supporting correct behavior at dynamically sensitive trajectory regions should become less plastic in later tasks.

Unlike reference-based consolidation methods, the proposed Dynamics-Risk Adaptive Plasticity (\method) does not penalize parameter displacement from a previously stored checkpoint. Instead, similar to CLASSP \cite{ludwig2024classp}, it uses accumulated historical importance to directly regulate future plasticity. Protection is therefore active from the first update of each subsequent learning stage and does not require storing reference copies of the adaptable parameters.

This work contributes a local policy-independent dynamics-risk estimator derived from expert state-action trajectories. The resulting risk scores weight the historical importance of adaptable tensors, which is then used to regulate their future plasticity through tensor-specific AdamW learning rates. This mechanism suppresses the actual optimizer step rather than merely scaling raw gradients.

The remainder of this paper is organized as follows. Section~\ref{sec:related} reviews related work on continual learning for vision-language-action models and on importance-based plasticity control. Section~\ref{sec:evaluation_protocol} describes the sequential LIBERO evaluation protocol. Section~\ref{sec:dynamics_risk_motivation} motivates the dynamics-risk signal from the compounding-error perspective of imitation learning, and Section~\ref{sec:problem} formalizes the replay-free continual learning problem. Section~\ref{sec:method} presents the DRAP method, including the KappaTune-based subnetwork selection, the local dynamics-risk estimator, the risk-weighted tensor importance, and the tensor-specific learning-rate control. Section~\ref{sec:experimental_results} reports the experimental results, including ablations of KappaTune selection and of dynamics-risk weighting. Finally, Section~\ref{sec:conclusion} concludes and outlines directions for future work.

\section{Related Work}
\label{sec:related}

Recent continual-learning methods for vision-language-action models primarily rely on architectural modularity, routing, replay, or parameter regularization. CLARE~\cite{romer2026clare} autonomously expands and routes lightweight adapters, allowing new manipulation tasks to be learned without requiring task identifiers at deployment. Stellar VLA~\cite{wu2026stellar} instead models task and skill relationships in a learned knowledge space and uses knowledge-guided expert routing, with limited replay, to reduce interference between sequentially learned behaviors. These approaches obtain stability by separating or routing task-specific capacity, whereas \method{} retains a fixed model architecture and controls the plasticity of an existing trainable subnetwork.

Among replay-free regularization methods, EWC~\cite{kirkpatrick2017ewc}, Synaptic Intelligence~\cite{zenke2017si}, Memory Aware Synapses~\cite{aljundi2018mas}, and RWalk~\cite{chaudhry2018rwalk} estimate the historical importance of model parameters and restrict subsequent changes to important parameters. EWC and RWalk employ reference-based quadratic penalties, while Synaptic Intelligence accumulates optimization-path contributions and Memory Aware Synapses measures the sensitivity of the learned output function. Despite their different motivations, these importance estimators have been shown to be closely related to Fisher-information-based quantities~\cite{benzing2022unifying}. DRAP differs by weighting sample-level action-gradient responsibility according to the estimated physical consequence of errors along the demonstrated trajectory.

DRAP is most closely related to continual-learning methods that regulate future parameter plasticity according to historical importance. CLASSP~\cite{ludwig2024classp} suppresses future updates using accumulated gradient magnitudes and optionally applies a gradient threshold to promote sparse adaptation. Bayesian approaches such as uncertainty-guided continual learning~\cite{ebrahimi2020ucb,ahn2019ucl} similarly use parameter uncertainty to determine which model components should remain stable and which should retain greater plasticity. DRAP follows this general principle of importance-dependent plasticity, but replaces generic update history or posterior uncertainty with dynamics-risk-weighted, sample-level behavioral responsibility. The accumulated importance is defined at tensor level and directly scales tensor-specific AdamW learning rates. Thus, DRAP is inspired by CLASSP's adjustment-suppression principle, but uses a different importance signal and optimizer interface.

Reference-based methods adopt a different protection mechanism. Elastic Weight Consolidation (EWC)~\cite{kirkpatrick2017ewc} estimates parameter importance using a diagonal Fisher approximation and penalizes displacement from a previous optimum. In contrast, DRAP does not store reference parameters or apply a displacement penalty. Historical importance regulates plasticity immediately from the first update of each subsequent learning stage.

Liu et al.~\cite{liu2026pretrained} investigate continual learning with large pretrained VLAs by sequentially training across the ten tasks of each LIBERO suite. Their main finding is that pretrained models such as $\pi_0$ and GR00T N1.5 can achieve little forgetting when combined with experience replay, and retain comparatively low forgetting even when the replay buffer contains only a small fraction of the previous-task data. However, their zero-replay Sequential and EWC baselines on LIBERO-Object and LIBERO-10 still exhibit substantial negative backward transfer, indicating that pretraining alone does not prevent forgetting when earlier demonstrations are unavailable.

Their protocol differs from ours in three respects. First, they train a separate continual model for each LIBERO suite and learn the ten tasks within that suite sequentially, whereas our protocol updates one checkpoint across complete suites in the order Spatial, Object, and Goal. Second, their strongest retention results store transitions from previous tasks, while DRAP is evaluated without storing any previous-suite demonstrations. Third, they use different VLA architectures with LoRA-based or partially frozen fine-tuning, whereas we adapt a KappaTune-selected subnetwork of OpenVLA. The two settings therefore address complementary questions: their results show that pretrained VLAs exploit small replay buffers particularly well, whereas DRAP targets retention under a strictly replay-free sequence of suite-level shifts. Since the replay-free setting is already difficult in both protocols, and our sequence additionally requires retention across suites that differ in layout, object identity, and goal semantics, we consider it a demanding test of forgetting.

KappaTune~\cite{ludwig2025kappatune} uses the matrix condition number as a scale-invariant proxy for information encoding in neural tensors. Its central hypothesis is that tensors with larger condition numbers contain more strongly differentiated singular directions and are therefore more likely to encode specialized pretrained information. KappaTune preserves these tensors and concentrates adaptation on tensors with smaller condition numbers, which are treated as safer sources of plasticity. 

\section{Evaluation Protocol} 
\label{sec:evaluation_protocol}
LIBERO provides a benchmark for knowledge transfer and lifelong robot learning across Spatial, Object, Goal, and Long-Horizon suites \cite{liu2023libero}. LIBERO results depend strongly on the training protocol. OpenVLA-OFT reports a $97.1\%$ average success rate using four independently fine-tuned policies, one for each of the Spatial, Object, Goal, and Long-Horizon suites \cite{kim2025oft}. It also reports $96.8\%$ for a single policy jointly trained on data from all four suites. These settings evaluate suite-specific or joint multi-suite acquisition. This work instead evaluates sequential continual learning without replay. A single checkpoint is trained successively on Spatial, Object and Goal, with only the current suite available at each stage. After every stage, the checkpoint is evaluated on all suites observed so far. The resulting lower-triangular success matrix therefore measures both new-suite acquisition and backward retention under closed-loop control.

As shown by the standard sequential-training baseline, this protocol presents a challenging catastrophic-forgetting setting, with performance on earlier suites degrading substantially after adaptation to subsequent suites.

\section{Compounding Errors and Dynamics-Risk Motivation}
\label{sec:dynamics_risk_motivation}
Behavior cloning is affected by covariate shift because policy-generated actions modify future states and observations \cite{ross2010efficient,ross2011dagger}. A small action error can move the robot away from the demonstrated trajectory, expose the policy to unfamiliar observations, and increase the probability of subsequent errors.

In physical AI, these deviations can compound over a closed-loop rollout and cause task failure even when the per step prediction error remains small. DRAP addresses this problem by assigning greater importance to demonstration samples at which the local physical dynamics are more likely to amplify a state perturbation. The risk estimator is inspired by Lyapunov's indirect method, which studies nonlinear systems locally through linearization \cite{khalil2002nonlinear}, and by contraction analysis, which examines the local convergence or divergence of neighboring trajectories \cite{lohmiller1998contraction}. DRAP uses this local differential perspective to rank demonstration samples according to their estimated one-step perturbation amplification.

The resulting dynamics-risk score is a relative sensitivity indicator, its purpose is to identify trajectory regions where preserving accurate policy behavior may be particularly important because small deviations can have larger downstream physical consequences.

\section{Problem Formulation}
\label{sec:problem}
Tasks arrive sequentially as $\mathcal{D}_1,\ldots,\mathcal{D}_K$. Each task dataset consists of expert transitions:
\begin{equation}
(o_t,s_t,a_t,s_{t+1},e_t),
\end{equation}
where $o_t$ is the visual-language observation, $s_t\in\mathbb{R}^{d_s}$ is a logged robot state, $a_t\in\mathbb{R}^{d_a}$ is the expert action, and $e_t$ identifies the episode. The VLA policy is $\pi_\theta(a\mid o)$ with parameters $\theta$.

A fixed set of trainable tensors, selected using KappaTune, is denoted by $\mathcal{T}$. For tensor $T\in\mathcal{T}$, $I_T^{(k)}\ge 0$ denotes cumulative importance after task $k$. The objective is to learn task $k+1$ while regulating each tensor's plasticity according to all previous task importance, without storing previous demonstrations.

\section{Method}
\label{sec:method}
\subsection{KappaTune selection}
For each eligible matrix $W$, \kappatune{} computes a regularized condition number
\begin{equation}
    \kappa(W)=\frac{\sigma_{\max}(W)}{\sigma_{\min}(W)+\eps_\kappa}.
\end{equation}
Within each architectural group (vision, language and action head), a fixed budget of low-$\kappa$ tensors is selected from the pretrained checkpoint. Selection is computed once and reused throughout the task sequence. Parameters outside $\mathcal{T}$ remain frozen. Therefore, \kappatune{} defines where adaptation may occur, while \method{} determines how much future plasticity each selected tensor retains.

\subsection{Local linear dynamics identification}
\begin{figure*}[t] 
   \centering
       \includegraphics[ width=0.72\textwidth, keepaspectratio ]{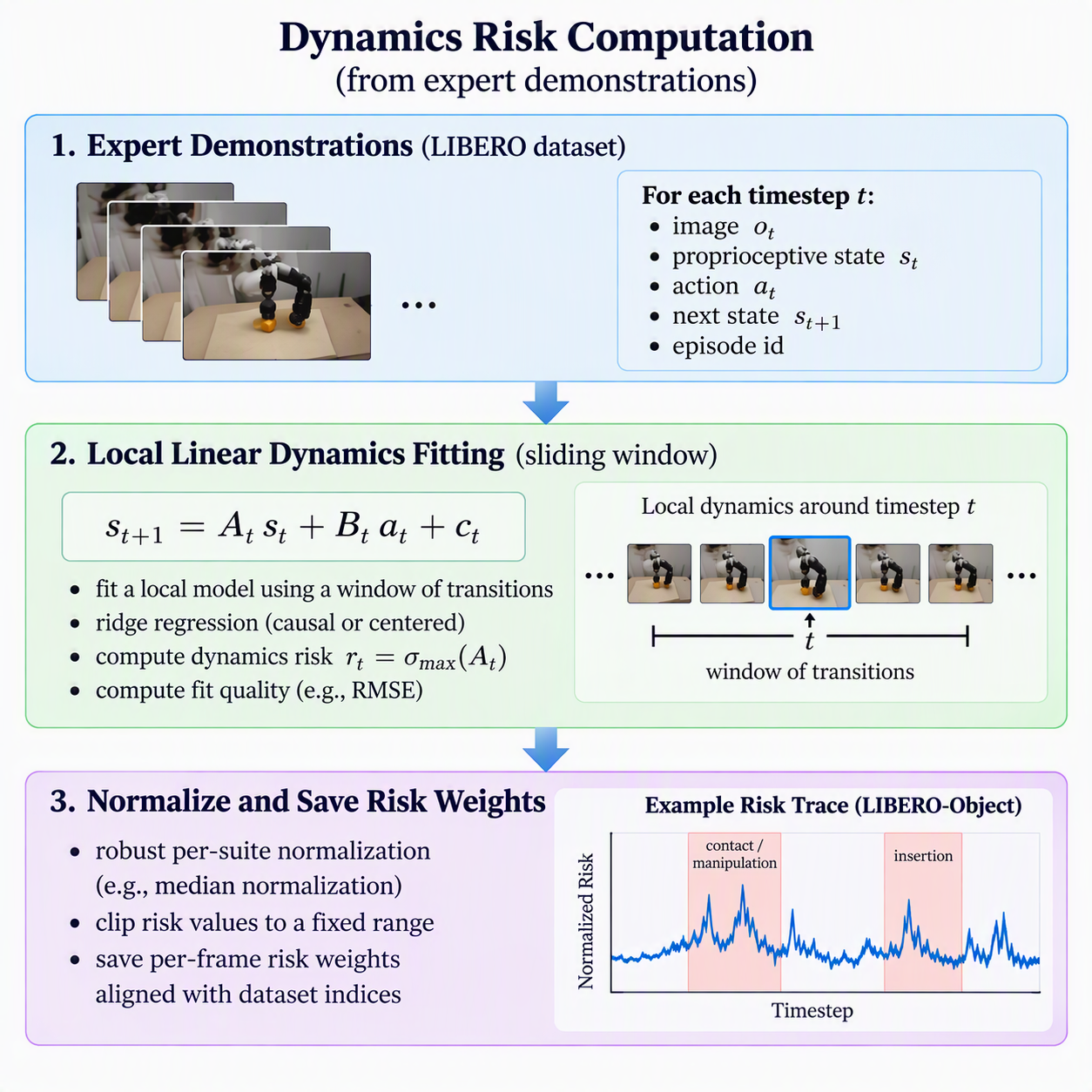}
			  \caption{ Dynamics-risk computation from expert demonstrations. Local linear dynamics are estimated over sliding windows of logged state-action transitions. The largest singular value $\sigma_{\max}(A_t)$ defines a local one-step perturbation-amplification score, which is robustly normalized and stored as a sample-level risk weight aligned with the training dataset. }
   \label{risk_computation}
\end{figure*}

State and action dimensions are robustly standardized using suite-level statistics. For a sliding window $\mathcal{W}_t$ contained within one episode, ridge regression estimates
\begin{equation}
    \label{linear}
    \tilde s_{j+1}=A_t\tilde s_j+B_t\tilde a_j+c_t+\epsilon_j,
    \qquad j\in\mathcal{W}_t,
\end{equation}
where $A_t\in\mathbb{R}^{d_s\times d_s}$ is the local state-transition matrix, $B_t\in\mathbb{R}^{d_s\times d_a}$ is the local control matrix, $c_t$ is an intercept, and $\epsilon_j$ is the residual. The model is local and trajectory-conditioned; it is not a global linear model of manipulation dynamics. See Figure \ref{risk_computation}.

\subsection{Lyapunov-inspired dynamics risk}

For a small state perturbation under the locally fixed logged action,
\begin{equation}
    \delta s_{t+1}\approx A_t\delta s_t,
\end{equation}
define the quadratic perturbation energy:
\begin{equation}
    V_s(\delta s)=\delta s^\top\delta s.
\end{equation}
Then the following upper-bound holds:
\begin{align}
    V_s(\delta s_{t+1})
    &=\delta s_t^\top A_t^\top A_t\delta s_t \\
    &\le \sigma_{\max}^2(A_t)V_s(\delta s_t).
\end{align}
Accordingly, we define the raw one-step dynamics sensitivity:
\begin{equation}
    \rho_t=\sigma_{\max}(A_t).
\end{equation}
Let $\mathcal{D}_k$ denote the set of demonstration samples in suite $k$. The corresponding training weight $r_t$ is obtained by normalizing $\rho_t$ with respect to the median raw dynamics sensitivity over $\mathcal{D}_k$ and clipping the result:
\begin{equation}
    \label{rt}
    r_t=\operatorname{clip}\left(
    \frac{\rho_t}{\operatorname{median}_{j\in\mathcal{D}_k}(\rho_j)+\eps_r},
    c_{\min},c_{\max}\right).
\end{equation}
where $\varepsilon_r>0$ ensures numerical stability, and $c_{\min}$ and $c_{\max}$ denote the lower and upper clipping bounds, respectively\footnote{For this specific experiment, the limits are arbitrarily set as $\left[c_{\min}, c_{\max}\right]=\left[0.25, 4\right]$.}.

This dynamics-risk score does not constitute a formal proof or certificate of closed-loop stability. Even if the policy-feedback contribution were included, the resulting closed-loop Jacobian would remain a first-order local approximation of the nonlinear robot--policy system around the expert trajectory. To clarify this limitation, consider the local dynamics model in \eqref{linear}, where $A_t$ describes the sensitivity of the standardized next state $\tilde{s}_{t+1}$ to the standardized current state $\tilde{s}_t$, $B_t$ describes the sensitivity of $\tilde{s}_{t+1}$ to the standardized action $\tilde{a}_t$, and $c_t$ is an affine offset. Let $o(\tilde{s})$ denote the observation generated from the standardized state, and let $\tilde{\pi}_\theta$ denote the policy expressed in standardized action coordinates. The local sensitivity of the policy action to the state is:
\begin{equation}
   J_{\pi,t} = \left. \frac{\partial \tilde{\pi}_\theta\!\left(o(\tilde{s})\right)} {\partial \tilde{s}} \right|_{\tilde{s}=\tilde{s}_t}.
\end{equation}

The corresponding first-order closed-loop Jacobian is:

\begin{equation}
   J_{\mathrm{CL},t} \approx A_t + B_tJ_{\pi,t}.
\end{equation}

The raw dynamics-risk score $\rho_t=\sigma_{\max}(A_t)$ captures only the direct state-sensitivity term $A_t$ and does not include the policy-feedback term $B_tJ_{\pi,t}$. However, even $J_{\mathrm{CL},t}$ would describe perturbation evolution only locally, near the linearization point and within the fitted operating regime. It would not account for higher-order nonlinearities, contact-mode transitions, observation aliasing, local-model estimation error, or trajectories that leave the neighborhood of the expert demonstration. A formal stability analysis would require additional conditions, such as a valid Lyapunov function over a specified region and a corresponding decrease condition along the actual nonlinear closed-loop dynamics.

Accordingly, the normalized score $r_t$ is interpreted only as a relative, policy-independent indicator of potential one-step physical perturbation amplification along the expert trajectory.

\subsection{Sample-Level Risk-Weighted Tensor Importance}
\label{subsec:frame_importance}

After completing task $k$, tensor importance is estimated from a deterministic subset of $N_k$ demonstration samples. A demonstration sample $i$ is a training example containing the policy observation $o_i$, the corresponding expert action $a_i$, and the aligned normalized dynamics-risk score $r_i$. The state-action transition used to estimate $r_i$ is associated with the same time step.

The importance-estimation loader uses a batch size of one. Consequently, the gradient produced by each sample is squared before being combined with the contributions of other samples. This prevents gradients associated with different samples, episodes, task phases, or action directions from canceling one another before the importance is calculated.

In the evaluated VLA, the continuous expert action $a_i$ is represented by a short sequence of discrete action tokens. Let $\mathcal{Q}_i$ denote the set of output-sequence positions associated with the expert action for sample $i$. An action-token position in the model output sequence represents one component of the discretized expert action. Positions corresponding to image patches, language instructions, prompt tokens, padding tokens, or other non-action content are excluded from $\mathcal{Q}_i$.

The sample-level imitation loss is defined as
\begin{equation}
    \ell_i(\theta)
    =
    \frac{1}{|\mathcal{Q}_i|}
    \sum_{q \in \mathcal{Q}_i}
    \ell_{i,q}(\theta),
    \label{eq:frame_loss}
\end{equation}
where $\ell_{i,q}(\theta)$ is the cross-entropy loss between the target action token and the token distribution predicted at position $q$, and $\theta$ denotes the model parameters.

Let $T \in \mathcal{T}$ be an adaptable parameter tensor, where $\mathcal{T}$ denotes the set of tensors selected via KappaTune for continual training. Let $n_T$ denote the number of scalar parameters contained in $T$. The gradient $\nabla_T \ell_i(\theta)$ has the same shape as $T$ and measures the sensitivity of the sample-level action loss to changes in that tensor.

Since $r_i$ is precomputed from the demonstration trajectory and is independent of the model parameters,
\begin{align}
    \left\|
        \nabla_T\left(r_i\ell_i(\theta)\right)
    \right\|_F^2
    &=
    \left\|
        r_i\nabla_T\ell_i(\theta)
    \right\|_F^2
    \nonumber\\
    &=
    r_i^2
    \left\|
        \nabla_T\ell_i(\theta)
    \right\|_F^2,
    \label{eq:implemented_risk_gradient}
\end{align}
where $\|\cdot\|_F$ denotes the Frobenius norm. Accordingly, the raw new-task importance of tensor $T$ after task $k$ is
\begin{equation}
    \widehat{I}_{T,\mathrm{new}}^{(k)}
    =
    \frac{1}{N_k}
    \sum_{i=1}^{N_k}
    r_i^2
    \frac{
        \left\|
            \nabla_T\ell_i(\theta)
        \right\|_F^2
    }{
        n_T
    }.
    \label{eq:importance}
\end{equation}
The quantity $\|\nabla_T\ell_i(\theta)\|_F^2/n_T$ is the mean squared gradient over all scalar elements of tensor $T$. Division by $n_T$ prevents larger tensors from receiving greater importance solely because they contain more parameters.

The estimator therefore contains three distinct reductions. First, the cross-entropy losses are averaged over the supervised action-token positions of each sample through \eqref{eq:frame_loss}. Second, the squared gradient is averaged over the $n_T$ scalar elements of tensor $T$. Third, the resulting risk-weighted contributions are averaged over the $N_k$ demonstration samples. The estimator produces one scalar raw importance coefficient for each adaptable tensor.

To express the normalization applied to both the new-task and cumulative importance, we define the normalization of a nonnegative tensor-level quantity $\{x_T\}_{T\in\mathcal{T}}$ as: 
\begin{equation}
\mathcal{N}(x)_T = \frac{x_T} {\displaystyle \operatorname{median}_{U\in\mathcal{T}:x_U>0}(x_U) +\varepsilon}.
\label{eq:positive_median_normalization}
\end{equation}
where $\varepsilon>0$ is a small constant that ensures numerical stability.

Accordingly, the new-task importance is normalized before being added to the cumulative memory:
\begin{equation}
\widetilde{I}_{T,\mathrm{new}}^{(k)} = \mathcal{N}\!\left( \widehat{I}_{\mathrm{new}}^{(k)} \right)_T.
\label{eq:new_task_importance_normalization}
\end{equation}
This normalization removes the task-dependent global scale while preserving the relative importance distribution across tensors.

The normalized new-task contribution is then merged into the cumulative importance memory:
\begin{equation}
    I_T^{(k)}
    =
    \gamma I_T^{(k-1)}
    +
    \widetilde{I}_{T,\mathrm{new}}^{(k)},
    \label{eq:importance_merge}
\end{equation}
where $I_T^{(k-1)}$ contains the importance accumulated from previous tasks and $\gamma \in (0,1]$ controls the retention of this historical importance. The default setting $\gamma=1$ preserves contributions from all earlier tasks.

Before training task $k+1$, the cumulative importance is normalized using the operator \eqref{eq:positive_median_normalization}:
\begin{equation}
\overline{I}_T^{(k)} = \mathcal{N}\!\left(I^{(k)}\right)_T.
\label{eq:cumulative_importance_normalization}
\end{equation}

DRAP converts this normalized cumulative importance into the tensor-specific plasticity multiplier
\begin{equation}
    s_T^{(k)}
    =
    \left(
        1+\beta\overline{I}_T^{(k)}
    \right)^{-1/p},
    \label{eq:drap_suppression}
\end{equation}
where $\beta\geq0$ controls the suppression strength and $p>0$ controls its curvature. The effective AdamW learning rate assigned to tensor $T$ during task $k+1$ is
\begin{equation}
    \eta_T^{(k+1)}
    =
    \eta s_T^{(k)},
    \label{eq:tensor_learning_rate}
\end{equation}
where $\eta$ is the base learning rate.

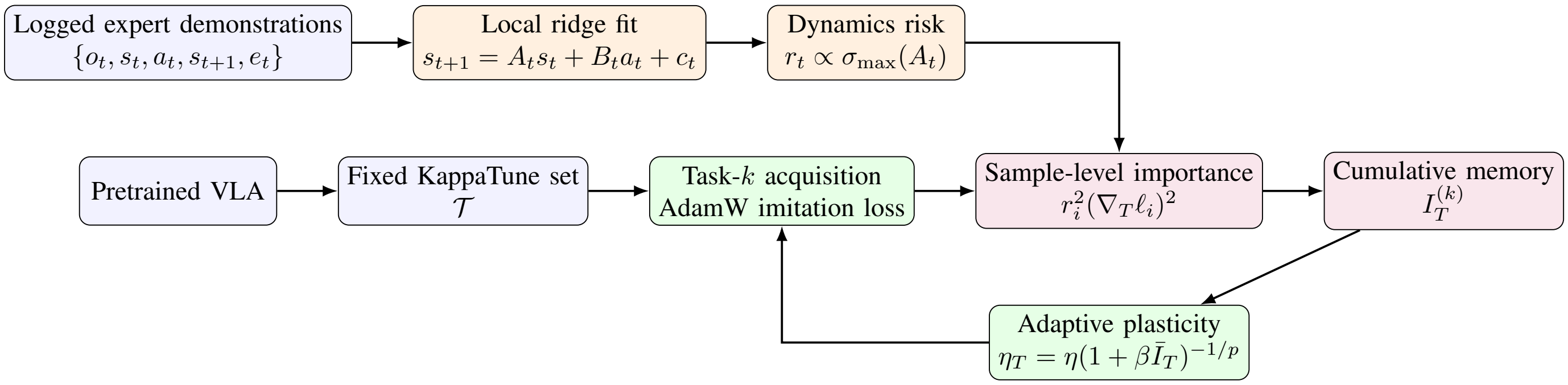
\begin{figure*}[t]
\centering
\resizebox{\textwidth}{!}{%
\begin{tikzpicture}[
node distance=8mm and 8mm,
box/.style={draw, rounded corners, align=center, minimum height=9mm, minimum width=26mm, fill=blue!5},
risk/.style={box, fill=orange!12},
train/.style={box, fill=green!10},
mem/.style={box, fill=purple!10},
arrow/.style={-{Latex[length=2mm]}, thick}
]
\node[box] (demo) {Logged expert demonstrations\\$\{o_t,s_t,a_t,s_{t+1},e_t\}$};
\node[risk, right=of demo] (fit) {Local ridge fit\\$s_{t+1}=A_ts_t+B_ta_t+c_t$};
\node[risk, right=of fit] (risk) {Dynamics risk\\$r_t\propto\sigma_{\max}(A_t)$};

\node[box, below=10mm of demo] (vla) {Pretrained VLA};
\node[box, right=of vla] (kt) {Fixed \kappatune{} set\\$\mathcal{T}$};
\node[train, right=of kt] (task) {Task-$k$ acquisition\\AdamW imitation loss};
\node[mem, right=of task] (importance) {Sample-level importance\\$r_i^2(\nabla_T\ell_i)^2$};
\node[mem, right=of importance] (memory) {Cumulative memory\\$I_T^{(k)}$};
\node[train, below=10mm of importance] (scale) {Adaptive plasticity\\$\eta_T=\eta(1+\beta\bar I_T)^{-1/p}$};

\draw[arrow] (demo)--(fit);
\draw[arrow] (fit)--(risk);
\draw[arrow] (vla)--(kt);
\draw[arrow] (kt)--(task);
\draw[arrow] (risk.east) -| (importance.north);
\draw[arrow] (task)--(importance);
\draw[arrow] (importance)--(memory);
\draw[arrow] (memory)--(scale);
\draw[arrow] (scale)-|(task);
\end{tikzpicture}%
}
\caption{Overview of \method. Expert trajectories provide a local dynamics-risk signal. Sample-level action gradients identify tensor responsibility at risky samples. Cumulative importance directly scales future AdamW learning rates; no previous tensor reference or displacement penalty is required.}
\label{fig:overview}
\end{figure*}

Thus, $I_T^{(k)}$ directly regulates the future plasticity of tensor $T$: tensors with larger historical importance receive smaller AdamW steps from the beginning of the next training stage, whereas tensors with lower importance remain comparatively plastic. Algorithm~\ref{alg:drap} summarizes the method, and Figure \ref{fig:overview} illustrates its overall workflow.

\begin{algorithm}[t]
\caption{KappaTune + DRAP}
\label{alg:drap}
\begin{algorithmic}[1]
\Require Pretrained policy $\theta_0$, task stream $\{\mathcal{D}_k\}_{k=1}^K$
\State Select fixed trainable tensor set $\mathcal{T}$ with \kappatune
\State Initialize $I_T^{(0)}\gets0$ for all $T\in\mathcal{T}$
\For{$k=1,\ldots,K$}
		\State Compute $\overline I_T^{(k-1)}$ by \eqref{eq:cumulative_importance_normalization} and $s_T^{(k-1)}$ by \eqref{eq:drap_suppression}
    \State Create AdamW groups with $\eta_T=\eta s_T^{(k-1)}$
    \For{each optimization step on $\mathcal{D}_k$}
        \State Compute imitation loss and accumulated gradient
        \State Apply AdamW update using tensor-specific $\eta_T$
    \EndFor
		\State Compute $\widehat I_{T,\mathrm{new}}^{(k)}$ by \eqref{eq:importance} and $\widetilde I_{T,\mathrm{new}}^{(k)}$ by \eqref{eq:new_task_importance_normalization}
    \State Merge cumulative importance by \eqref{eq:importance_merge}
    \State Save checkpoint, fixed tensor set $\mathcal{T}$
		\State Save cumulative importance $\{I_T^{(k)}\}_{T\in\mathcal{T}}$
\EndFor
\end{algorithmic}
\end{algorithm}

\begin{table*}[t]
\centering
\caption{Comparison of CLASSP, EWC, and the proposed DRAP.}
\label{tab:comparison}
\begin{tabular}{p{0.18\textwidth}p{0.25\textwidth}p{0.25\textwidth}p{0.25\textwidth}}
\toprule
Dimension & CLASSP & EWC & Proposed DRAP \\
\midrule
Historical signal
& Powered gradient/update accumulation
& Diagonal Fisher information estimated from previous-task gradients
& Dynamics-risk-weighted sample-level gradients \\
\cmidrule(lr){1-4}

Physical grounding
& None
& None
& Local trajectory-amplification risk \\
\cmidrule(lr){1-4}

Protection mechanism
& Update denominator and gradient threshold
& Fisher-weighted quadratic displacement penalty
& Tensor-specific AdamW learning rate \\
\cmidrule(lr){1-4}

Reference weights
& Not required
& Required, $\theta^\star$
& Not required \\
\cmidrule(lr){1-4}

Onset of protection
& Immediate
& Zero at the new-task initial point; increases with displacement from $\theta^\star$
& Immediate \\
\cmidrule(lr){1-4}

Optimizer
& Custom CLASSP optimizer
& AdamW plus EWC regularization loss
& AdamW with per-tensor learning-rate groups \\
\cmidrule(lr){1-4}

Threshold
& Core sparsity mechanism
& None
& None \\
\cmidrule(lr){1-4}

Importance granularity
& Usually parameter-level accumulated adjustment
& Parameter-level diagonal Fisher
& Tensor-level cumulative importance from sample-level gradients \\
\cmidrule(lr){1-4}

Primary risk
& Excessive suppression can block new-task acquisition
& Inaccurate Fisher estimates or excessive regularization can restrict plasticity
& Important tensors can still accumulate drift through repeated reduced updates \\
\cmidrule(lr){1-4}

Replay
& Not required
& Not required after Fisher estimation
& Not required after importance estimation \\
\cmidrule(lr){1-4}

Stored continual-learning state
& Accumulated adjustment statistics
& Diagonal Fisher and reference parameters
& Cumulative tensor importance only \\

\bottomrule
\end{tabular}
\end{table*}

\section{Experimental Results}
\label{sec:experimental_results}

We evaluate five sequential-training configurations on the task order Spatial $\rightarrow$ Object $\rightarrow$ Goal: standard sequential fine-tuning, KappaTune+EWC, KappaTune+DRAP, KappaTune+DRAP with uniform risk, and DRAP without KappaTune. The uniform-risk ablation sets $r_i=1$ for every sample while retaining sample-level gradient importance, cumulative importance memory, and tensor-specific AdamW learning rates\footnote{The source code is available at https://github.com/oswaldoludwig/DRAP-Dynamics-Risk-Adaptive-Plasticity}.

Each checkpoint is evaluated after every training stage on all suites observed up to that point. To reduce finite-rollout variability, this experiment pools 6 Gym evaluation runs, each comprising 50 episodes. These repeated evaluations quantify rollout variability for fixed checkpoints.

\subsection{Evaluation Metrics}

Let $S_{k,j}$ denote the pooled success rate on suite $j$ after training stage $k$. The average success over all suites observed by stage $k$ is
\begin{equation}
    \operatorname{AvgSuccess}_k
    =
    \frac{1}{k+1}
    \sum_{j=0}^{k}
    S_{k,j}.
    \label{eq:avg_success}
\end{equation}

For a previous suite $j<k$, forgetting is measured relative to its best earlier performance:
\begin{equation}
    F_{k,j}
    =
    \max_{m\in\{j,\ldots,k-1\}}
    S_{m,j}
    -
    S_{k,j}.
    \label{eq:forgetting}
\end{equation}
A positive value indicates forgetting.

Average forgetting after stage $k$ is
\begin{equation}
    \operatorname{AvgForget}_k
    =
    \frac{1}{k}
    \sum_{j=0}^{k-1}
    F_{k,j}.
    \label{eq:avg_forgetting}
\end{equation}

\subsection{Sequential Success and Forgetting Rates}

Table~\ref{tab:success_matrices} reports the pooled lower-triangular success matrix for each method. Each row corresponds to a training stage, and each column reports closed-loop success on an observed LIBERO suite.

\begin{table*}[t]
\centering
\caption{Sequential LIBERO success rates across repeated Gym evaluations.}
\label{tab:success_matrices}
\begin{tabular}{llccc}
\toprule
Method & Checkpoint & Spatial & Object & Goal \\
\midrule

\multirow{3}{*}{Standard fine-tuning}
& After Spatial & 0.77 & -- & -- \\
& After Object  & 0.11 & 0.83 & -- \\
& After Goal    & 0.00 & 0.07 & 0.71 \\
\cmidrule(lr){1-5}

\multirow{3}{*}{KappaTune+EWC}
& After Spatial & 0.76 & -- & -- \\
& After Object  & 0.41 & 0.82 & -- \\
& After Goal    & 0.03 & 0.30 & 0.64 \\
\cmidrule(lr){1-5}

\multirow{3}{*}{KappaTune+DRAP}
& After Spatial & 0.77 & -- & -- \\
& After Object  & \textbf{0.75} & 0.80 & -- \\
& After Goal    & \textbf{0.34} & \textbf{0.61}
& \textbf{0.71} \\
\cmidrule(lr){1-5}

\multirow{3}{*}{KappaTune+Uniform-risk DRAP}
& After Spatial & 0.78 & -- & -- \\
& After Object  & 0.64 & 0.80 & -- \\
& After Goal    & 0.29 & 0.57 & 0.66 \\
\cmidrule(lr){1-5}

\multirow{3}{*}{DRAP without KappaTune}
& After Spatial & 0.78 & -- & -- \\
& After Object  & 0.71 & 0.82 & -- \\
& After Goal    & 0.11 & 0.08 & 0.66 \\
\bottomrule
\end{tabular}
\end{table*}

Table~\ref{tab:summary_metrics} summarizes average success and forgetting after Object and Goal training.

\begin{table*}[t]
\centering
\caption{Average success and forgetting derived from the pooled sequential evaluation matrices. Forgetting is measured relative to the best earlier checkpoint for each previously learned suite.}
\label{tab:summary_metrics}
\begin{tabular}{lcccccc}
\toprule
& \multicolumn{2}{c}{After Object}
& \multicolumn{4}{c}{After Goal} \\
\cmidrule(lr){2-3}
\cmidrule(lr){4-7}
Method
& Avg. success
& Spatial forget.
& Avg. success
& Spatial forget.
& Object forget.
& Avg. forget. \\
\midrule

Standard fine-tuning
& 0.47 & 0.66
& 0.26 & 0.77 & 0.76 & 0.77 \\
\cmidrule(lr){1-7}

KappaTune+EWC
& 0.62 & 0.35
& 0.32 & 0.73 & 0.52 & 0.63 \\
\cmidrule(lr){1-7}

KappaTune+DRAP
& \textbf{0.77} & \textbf{0.02}
& \textbf{0.55} & \textbf{0.43} & \textbf{0.19}
& \textbf{0.31} \\
\cmidrule(lr){1-7}

KappaTune+Uniform-risk DRAP
& 0.72 & 0.14
& 0.51 & 0.49 & 0.23 & 0.36 \\
\cmidrule(lr){1-7}

DRAP without KappaTune
& 0.77 & 0.07
& 0.28 & 0.67 & 0.74 & 0.71 \\
\bottomrule
\end{tabular}
\end{table*}

Standard sequential fine-tuning exhibits strongly recency-dominated behavior. Each newly introduced suite is acquired successfully, while performance on earlier suites rapidly approaches zero.

KappaTune+EWC provides partial protection during the first task transition, but its retention deteriorates substantially after Goal adaptation. EWC preserves more of the immediately preceding Object suite than standard fine-tuning, but offers little protection for the earliest Spatial suite. This result indicates that Fisher-weighted anchoring improves short-term retention but remains insufficient over the complete evaluated sequence.

KappaTune+DRAP achieves the strongest overall stability-plasticity balance. The resulting success matrix yields both the highest average success and the lowest average forgetting after Object and Goal training.

\subsection{Ablation of KappaTune Selection}

DRAP without KappaTune initially exhibits strong plasticity and performs competitively through the Object stage. However, after Goal adaptation, performance on both previous suites approaches the standard fine-tuning baseline. The broader adaptable parameter space permits effective short-term acquisition but exposes more pathways for destructive interference. The experiment indicates that KappaTune and DRAP provide complementary forms of protection. KappaTune restricts adaptation to a structurally selected subnetwork, while DRAP allocates plasticity within that subnetwork according to accumulated historical responsibility.

\subsection{Ablation of Dynamics-Risk Weighting}

The uniform-risk ablation sets $r_i=1$ for every sample while preserving all other elements of DRAP, including sample-level squared-gradient importance, cumulative importance memory, tensor-wise normalization, and tensor-specific AdamW learning rates. It therefore isolates the contribution of the demonstration-derived physical-risk signal from the broader adaptive-plasticity mechanism.

Uniform-risk DRAP retains much of the improvement over standard fine-tuning and EWC, showing that historical tensor importance and tensor-specific learning rates account for a substantial part of DRAP's performance. Nevertheless, dynamics-risk DRAP provides better retention of both Spatial and Object. New-suite acquisition remains comparable, indicating that the additional retention does not result from simply preventing the model from learning the current suite.

These initial results suggest that physics-informed plasticity control is a promising direction and motivate extending the dynamics-risk estimator beyond the state-transition term $A_t$ to include control-path sensitivity, multi-step perturbation gains, and contact-aware or learned local dynamics, potentially yielding a stronger stability-plasticity balance.

\section{Conclusion}
\label{sec:conclusion}

This work takes a first step toward physics-informed plasticity control by introducing DRAP, a replay-free method for continual learning in physical AI. DRAP estimates tensor importance from sample-level action gradients weighted by local trajectory-amplification risk and uses the accumulated importance to regulate tensor-specific AdamW learning rates during subsequent adaptation. Intuitively, DRAP suppresses future updates to tensors that were important for predicting expert actions at samples where errors are more strongly amplified by the physical dynamics.

Sequential LIBERO experiments showed that standard fine-tuning suffers severe catastrophic forgetting, while the physically ungrounded method, EWC, provides only partial retention. DRAP achieved the strongest stability-plasticity balance when combined with KappaTune. KappaTune restricts adaptation to a structurally selected subnetwork, while DRAP distributes plasticity within that subnetwork according to historical, dynamics-risk-weighted responsibility. The ablation without KappaTune confirmed that broader plasticity improves initial acquisition but increases later interference.

Sequential LIBERO experiments showed that standard fine-tuning suffers severe catastrophic forgetting, while EWC provides only partial retention. KappaTune+DRAP achieved the strongest stability-plasticity balance. KappaTune restricts adaptation to a selected subnetwork, while DRAP regulates plasticity within that subnetwork according to accumulated historical importance. The ablation without KappaTune showed that a broader adaptable parameter space supports strong initial acquisition but substantially increases later interference. The uniform-risk ablation showed that importance-dependent plasticity accounts for most of DRAP's benefit, while dynamics-risk weighting provides an additional improvement in retention without impairing new-suite acquisition.

These results support the hypothesis that continual learning in physical AI should consider both model sensitivity and the physical consequences of behavioral errors. Future work will improve the dynamics-risk estimator by incorporating the control pathway, online estimation, multi-step perturbation propagation, and contact-aware dynamics. It will also investigate the CLASSP-inspired update threshold and evaluate DRAP over longer continual-learning sequences.

\bibliographystyle{IEEEtran}
\bibliography{drap_references}

@article{ludwig2025kappatune,
  title={The Condition Number as a Scale-Invariant Proxy for Information Encoding in Neural Units},
  author={Ludwig, Oswaldo},
  journal={arXiv preprint arXiv:2506.16289},
  year={2025}
}

@article{kim2024openvla,
  title         = {{OpenVLA}: An Open-Source Vision-Language-Action Model},
  author        = {Kim, Moo Jin and others},
  journal       = {arXiv preprint arXiv:2406.09246},
  year          = {2024}
}

@inproceedings{kim2025oft,
  title         = {Fine-Tuning Vision-Language-Action Models: Optimizing Speed and Success},
  author        = {Kim, Moo Jin and Finn, Chelsea and Liang, Percy},
  booktitle     = {Robotics: Science and Systems},
  year          = {2025},
  eprint        = {2502.19645},
  archivePrefix = {arXiv},
  primaryClass  = {cs.RO}
}

@inproceedings{liu2023libero,
  title         = {{LIBERO}: Benchmarking Knowledge Transfer for Lifelong Robot Learning},
  author        = {Liu, Bo and others},
  booktitle     = {Advances in Neural Information Processing Systems},
  year          = {2023}
}

@article{romer2026clare,
  title         = {{CLARE}: Continual Learning for Vision-Language-Action Models via Autonomous Adapter Routing and Expansion},
  author        = {R{\"o}mer, Ralf and Zhang, Yi and Li, Yuming and Schoellig, Angela P.},
  journal       = {IEEE Robotics and Automation Letters},
  year          = {2026},
  doi           = {10.1109/LRA.2026.3693992}
}

@article{wu2026stellar,
  title         = {Continually Evolving Skill Knowledge in Vision Language Action Model},
  author        = {Wu, Yuxuan and Wang, Guangming and Yang, Zhiheng and Deng, Tianchen and Yao, Maoqing and Sheil, Brian and Wang, Hesheng},
  journal       = {arXiv preprint arXiv:2511.18085},
  year          = {2025}
}

@article{kirkpatrick2017ewc,
  title         = {Overcoming Catastrophic Forgetting in Neural Networks},
  author        = {Kirkpatrick, James and others},
  journal       = {Proceedings of the National Academy of Sciences},
  volume        = {114},
  number        = {13},
  pages         = {3521--3526},
  year          = {2017},
  doi           = {10.1073/pnas.1611835114}
}

@article{ludwig2024classp,
  title         = {{CLASSP}: A Biologically-Inspired Approach to Continual Learning through Adjustment Suppression and Sparsity Promotion},
  author        = {Ludwig, Oswaldo},
  journal       = {arXiv preprint arXiv:2405.09637},
  year          = {2024}
}

@inproceedings{ross2010efficient,
  title         = {Efficient Reductions for Imitation Learning},
  author        = {Ross, St{\'e}phane and Bagnell, Drew},
  booktitle     = {Proceedings of the Thirteenth International Conference on Artificial Intelligence and Statistics},
  year          = {2010}
}

@inproceedings{ross2011dagger,
  title         = {A Reduction of Imitation Learning and Structured Prediction to No-Regret Online Learning},
  author        = {Ross, St{\'e}phane and Gordon, Geoffrey J. and Bagnell, Drew},
  booktitle     = {Proceedings of the Fourteenth International Conference on Artificial Intelligence and Statistics},
  year          = {2011}
}

@article{lohmiller1998contraction,
  title         = {On Contraction Analysis for Non-linear Systems},
  author        = {Lohmiller, Winfried and Slotine, Jean-Jacques E.},
  journal       = {Automatica},
  volume        = {34},
  number        = {6},
  pages         = {683--696},
  year          = {1998},
  doi           = {10.1016/S0005-1098(98)00019-3}
}

@book{khalil2002nonlinear,
  title         = {Nonlinear Systems},
  author        = {Khalil, Hassan K.},
  edition       = {3},
  publisher     = {Prentice Hall},
  address       = {Upper Saddle River, NJ, USA},
  year          = {2002}
}

@inproceedings{zenke2017si,
  title         = {Continual Learning Through Synaptic Intelligence},
  author        = {Zenke, Friedemann and Poole, Ben and Ganguli, Surya},
  booktitle     = {Proceedings of the 34th International Conference on Machine Learning},
  series        = {Proceedings of Machine Learning Research},
  volume        = {70},
  pages         = {3987--3995},
  year          = {2017}
}

@inproceedings{aljundi2018mas,
  title         = {Memory Aware Synapses: Learning What (Not) to Forget},
  author        = {Aljundi, Rahaf and Babiloni, Francesca and Elhoseiny, Mohamed and Rohrbach, Marcus and Tuytelaars, Tinne},
  booktitle     = {Proceedings of the European Conference on Computer Vision},
  pages         = {139--154},
  year          = {2018}
}

@inproceedings{chaudhry2018rwalk,
  title         = {Riemannian Walk for Incremental Learning: Understanding Forgetting and Intransigence},
  author        = {Chaudhry, Arslan and Dokania, Puneet K. and Ajanthan, Thalaiyasingam and Torr, Philip H. S.},
  booktitle     = {Proceedings of the European Conference on Computer Vision},
  pages         = {532--547},
  year          = {2018}
}

@inproceedings{benzing2022unifying,
  title         = {Unifying Importance Based Regularisation Methods for Continual Learning},
  author        = {Benzing, Frederik},
  booktitle     = {Proceedings of the 25th International Conference on Artificial Intelligence and Statistics},
  series        = {Proceedings of Machine Learning Research},
  volume        = {151},
  pages         = {2372--2396},
  year          = {2022}
}

@inproceedings{ebrahimi2020ucb,
  title         = {Uncertainty-Guided Continual Learning with Bayesian Neural Networks},
  author        = {Ebrahimi, Sayna and Elhoseiny, Mohamed and Darrell, Trevor and Rohrbach, Marcus},
  booktitle     = {International Conference on Learning Representations},
  year          = {2020}
}

@inproceedings{ahn2019ucl,
  title         = {Uncertainty-Based Continual Learning with Adaptive Regularization},
  author        = {Ahn, Hongjoon and Cha, Sungmin and Lee, Donggyu and Moon, Taesup},
  booktitle     = {Advances in Neural Information Processing Systems},
  volume        = {32},
  year          = {2019}
}

@article{liu2026pretrained,
  title         = {Pretrained Vision-Language-Action Models are Surprisingly
                   Resistant to Forgetting in Continual Learning},
  author        = {Liu, Huihan and Kim, Changyeon and Liu, Bo and
                   Liu, Minghuan and Zhu, Yuke},
  journal       = {arXiv preprint arXiv:2603.03818},
  year          = {2026},
  doi           = {10.48550/arXiv.2603.03818}
}

\end{document}